\documentclass{article}
\usepackage{spconf,amsmath,graphicx}
\usepackage{bbding, makecell}
\usepackage{url}

\title{STUDENT DANGEROUS BEHAVIOR DETECTION IN SCHOOL}

\name{Huayi Zhou$^{\star}$ \qquad Fei Jiang$^{\dagger}$ \qquad Hongtao Lu$^{\star}$}
\address{$^{\star}$ Shanghai Jiao Tong University, sjtu\_zhy@sjtu.edu.cn, htlu@sjtu.edu.cn \\
    $^{\dagger}$ East China Normal University, fjiang@mail.ecnu.edu.cn }

\begin{document}
%
\maketitle
\begin{abstract}
Video surveillance systems have been installed to ensure the student safety in schools. However, discovering dangerous behaviors, such as fighting and falling down, usually depends on untimely human observations. In this paper, we focus on detecting dangerous behaviors of students automatically, which faces numerous challenges, such as insufficient datasets, confusing postures, keyframes detection and prompt response. To address these challenges, we first build a danger behavior dataset with locations and labels from surveillance videos, and transform action recognition of long videos to an object detection task that avoids keyframes detection. Then, we propose a novel end-to-end dangerous behavior detection method, named DangerDet, that combines multi-scale body features and keypoints-based pose features. We could improve the accuracy of behavior classification due to the highly correlation between pose and behavior. On our dataset, DangerDet achieves 71.0\% mAP with about 11 FPS. It keeps a better balance between the accuracy and time cost.
\end{abstract}

\section{Introduction}

\begin{figure}[htb]
\centering
\begin{minipage}[b]{0.63\linewidth}
\centerline{\includegraphics[width=5.5cm]{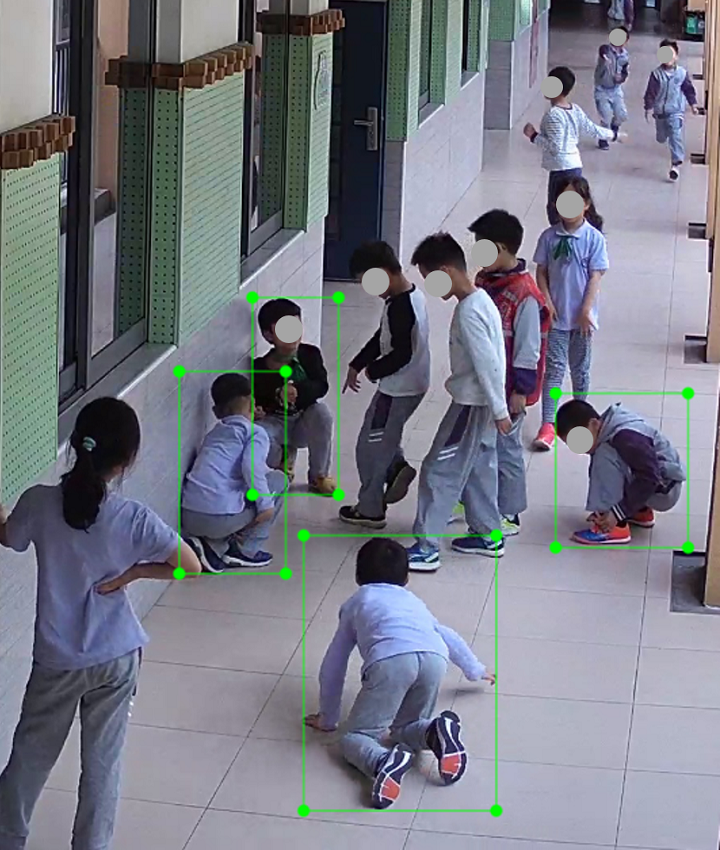}}
\centerline{(a)}\medskip
\end{minipage} 
\begin{minipage}[b]{0.36\linewidth}
\centerline{\includegraphics[width=3cm]{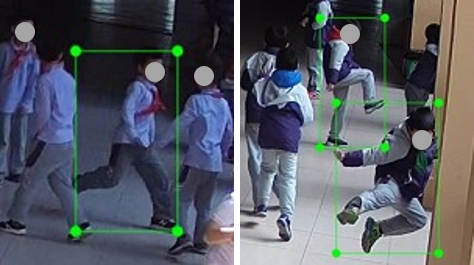}}
\centerline{(b) fight}\medskip
\centerline{\includegraphics[width=3cm]{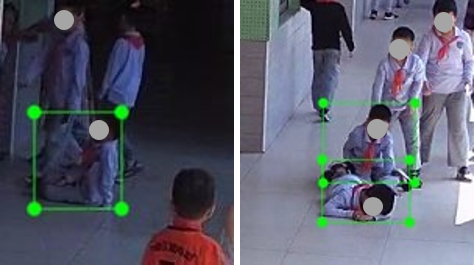}}
\centerline{(c) tumble}\medskip
\centerline{\includegraphics[width=3cm]{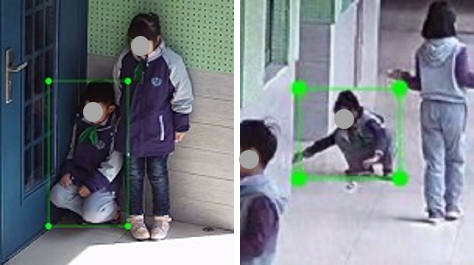}}
\centerline{(d) suqat}\medskip
\end{minipage}
\caption{Figure (a) shows the status of students in one corridor after class. Figure (b), (c) and (d) are annotation instances of three dangerous behaviors.}
\label{fig1}
\end{figure}

Recently, video surveillance systems are common to discover dangerous behaviors timely in school and shelter students from potential harm. Therefore, an effective and efficient monitoring alarm method for student dangerous behaviors is vital and urgent. However, student dangerous behaviors recognition has numerous challenges due to insufficient datasets, confusing postures, keyframes detection, and prompt response. In this paper, we focus on the primary school scene which is full of pupils who might behave dangerously because of their immaturity. Three typical potential dangerous behaviors, including fighting, tumbling and squatting, are automatically detected, as shown in Fig. \ref{fig1}.

Generally, dangerous behavior detection belongs to action recognition. The representative methods in this area include Two-stream convolutional networks \cite{simonyan2014two}, Temporal Segment Networks (TSN) \cite{wang2016temporal} and YOWO \cite{kopuklu2019you}. However, the common disadvantages of these approaches include the expensive computation and consecutive frames analysis, which are unfriendly to the practical applications. To satisfy the real applications, instead of applying SOTA yet cumbersome methods in the action recognition area, we propose a novel idea that combines lightweight keyframe feature extraction network and 2D human pose features to fast localize and more accurately recognize dangerous behaviors. Both of object detection and pose estimation have been used for action recognition independently, but rarely together. Several works with similar opinions to ours were proposed, such as UAV surveillance system \cite{penmetsa2014autonomous}, Drone Surveillance System (DSS) \cite{singh2018eye} and student behavior recognition system \cite{lin2021student}, which cannot be trained end-to-end.

\begin{figure*}[!t]
	\centering
	\includegraphics[width=0.9\textwidth]{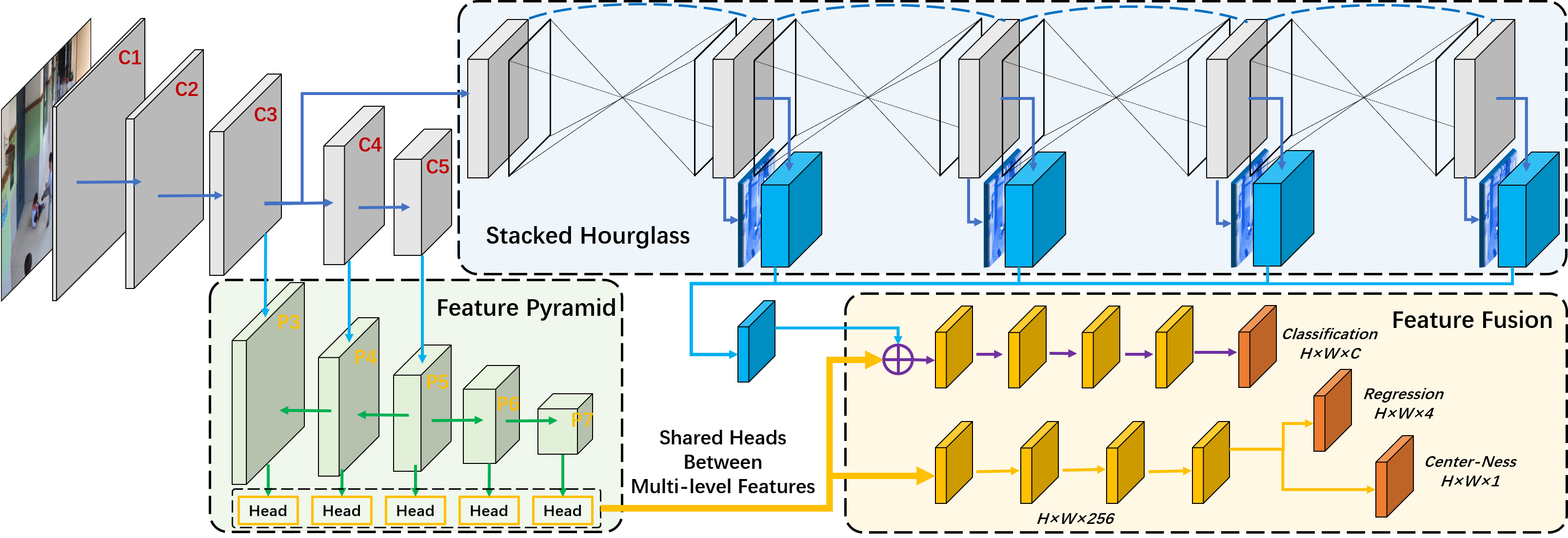}
	\caption{The network architecture of DangerDet. We use $\mathcal{C}_1 \sim \mathcal{C}_5$ to denote the feature maps from the FPN \cite{lin2017feature} based on ResNet \cite{he2016deep}. After $\mathcal{C}_3$ are four stacked hourglasses \cite{newell2016stacked} to extract pose features for better behavior classification. Feature levels $\mathcal{P}_3 \sim \mathcal{P}_7$ in feature pyramid are used for final prediction. Best viewed on screen.}
	\label{fig2}
\end{figure*}

Several researches have demonstrated the effectiveness and efficiency of specific behaviors recognition using object detection methods. Zheng et. al \cite{zheng2020intelligent} successfully applied improved Faster R-CNN \cite{ren2015faster} to the student behavior analysis, including hand-raising, standing and sleeping. After that, Zheng et. al \cite{zheng2020gesturedet} utilized lightweight network MobileNetV2-SSD \cite{sandler2018mobilenetv2} with multi-dimensional attention mechanisms to significantly reduce the computational cost. However, they ignore the pose features that are highly related to the behaviors. Meanwhile, EduSense \cite{ahuja2019edusense} adopted OpenPose \cite{cao2019openpose} for the behavior recognition of teachers and students in classrooms. Due to the highly occlusion among students, the real performances based on the 2D keypoints are obviously reduced. Compared with pose estimation methods, object detection algorithms for behavior recognition are less sensitive to the occlusion. Therefore, we propose a novel framework for the dangerous behavior recognition with object detection architecture appending pose estiamtion as the backbone.

Specifically, we firstly bulit a particular danger behavior dataset under school corridor scene. Then, we proposed the DangerDet for tackling dangerous behavior detection. The network architecture of DangerDet is shown in Fig. \ref{fig2}. We transform action recognition problem into behavior detection. The detection framework draws on the idea from FPN \cite{lin2017feature} and the single-stage anchor-free method FCOS \cite{tian2019fcos} for its good balance between accuracy and time-consuming. And in order to use the information of keypoints to improve the accuracy of behavior classification, the backbone network introduces the versatile stacked hourglass \cite{newell2016stacked} structure. In experiments, we verified the robustness and superiority of DangerDet compared with baseline on our self-made dataset. 

\section{Related Work}

In this section, we will briefly introduce the works related to deep feature extraction and keypoint heatmap generation involved in our DangerDet.

\subsection{Deep Feature Extraction}

The deep feature extraction based on convolutional neural network (CNN) and correlation strategies is an essential part of advanced object recognition algorithms \cite{ren2015faster, dai2016r, sandler2018mobilenetv2, lin2017feature, tian2019fcos}. One of the most important ideas to deal with the object scale variation is the feature pyramid strategy. This method is firstly proposed in FPN \cite{lin2017feature} which excavated hierarchical features from discretized image pyramid and made significant improvement in object recognition. Then, based on this foundation work, many similar pyramid structures with multi-scale lightweight feature maps and subtle skip connections architecture are invented. The effective feature pyramid in FCOS \cite{tian2019fcos} is a commonly used framework. For our DangerDet, we will adopt a homologous pyramid architecture for feature extraction to better and faster detect various dangerous behaviors. The legend is depicted in Fig. \ref{fig2}.

\subsection{Keypoint Heatmap Generation}

Here we introduce two representative pose estimation algorithms Stacked Hourglass \cite{newell2016stacked} and PifPaf \cite{kreiss2019pifpaf} which are conducive to our work.  The hourglass network structure is firstly proposed in \cite{newell2016stacked} to utilize multi-scale features produced by residual module in ResNet \cite{he2016deep} to recognize person pose. By stacking multiple such hourglasses blocks, we could reuse the information of whole-body joints to further improve the recognition accuracy. Generally, the four stacked hourglasses configuration is the most cost-effective and thus embedded into our architecture to generate keypoint heatmaps for enhancing our behavior detection task. As for PifPaf \cite{kreiss2019pifpaf}, it is an excellent open-source bottom-up multi-person pose estimation algorithm that outperforms previous methods under various scenes including low resolution, crowd and occlusion thanks to (i) the new composite field PAF encoding fine-grained information and (ii) the choice of Laplace loss for regressions which incorporates a notion of uncertainty. We directly use PifPaf to estimate the human pose of all the keyframes in our self-made dataset. The obtained keypoints will be used to generate keypoints heatmaps as the weak ground-truths to support the supervised training.

\section{Our Method}

In this section, we will first explain the network structure of our proposed method in detail. Then, the proposed feature aggregation strategy about multiple branches is illustrated. Finally, we will present the design of loss function about outputted features by multi-channel.

\subsection{Network Architecture}

The overall architecture of our DangerDet is the combination of both object detection and pose estimation. In Fig. \ref{fig2}, the input image shape is $(1152, 768, 3)$. After passing $\mathcal{C}_1$ and $\mathcal{C}_2$, the size of feature map is down-sampled 4$(2^2)$ times. Finally, the size of the feature map outputted by $\mathcal{C}_5$ is reduced to 32$(2^5)$ times. Based on blocks $\mathcal{C}_3 \sim \mathcal{C}_5$, we construct the feature pyramid $\mathcal{P}_3 \sim \mathcal{P}_7$ following FPN \cite{lin2017feature}. We share the heads between different feature levels of $\mathcal{P}_i (i\in[3,4,5,6,7])$ for multi-level prediction. To regress different size range among $\mathcal{P}_i$, we increase a trainable scalar $s_i$ to automatically adjust the base of the exponential function $exp(s_ix)$ for feature level $\mathcal{P}_i$. Besides, right after $\mathcal{C}_3$, we add four stacked hourglass modules connected in series. The intermediate supervision and symmetric distribution of capacity strategy in \cite{newell2016stacked} are maintained. These modules generate multiple keypoints heatmaps and pose features which are consistent with the output of $\mathcal{C}_3$ in size. Among them, the keypoints heatmap is used for supervised learning, and its dimension is $K$ which denotes the keypoints number. And pose features are abstracted  for subsequent behavior classification.

After shared heads, the final outputs have two pipelines and three branches similar to FCOS \cite{tian2019fcos}. The difference is that we have only three categories ($C=3$) in the \emph{Classification} branch. Moreover, we fuse all feature maps outputted by $\mathcal{P}_i$ with the combined pose feature map from the stacked hourglass module to obtain new augmented feature maps as inputs separately. For \emph{Regression} branch, it outputs a 4D vector $(l,t,r,b)$ encoding the location of a bounding box at each foreground pixel. The \emph{Center-Ness} branch depicts the normalized distance from the location to the center of the object that the location is responsible for. It is invented to down-weight the scores of bounding boxes far from the center of an object. Given the regression targets $l^*$, $t^*$, $r^*$, and $b^*$ for a location, it is defined in Eqn. \ref{eqn1} as below:

\begin{equation}\small
  CenterNess^*=\sqrt{\frac{min(l^*, r^*)}{max(l^*, r^*)} \times \frac{min(t^*, b^*)}{max(t^*, b^*)}}~
  \label{eqn1}
\end{equation}

\subsection{Feature Aggregation}

\begin{figure}[!t]
	\centering
	\begin{minipage}{0.58\linewidth}
	\leftline{\includegraphics[width=1\textwidth]{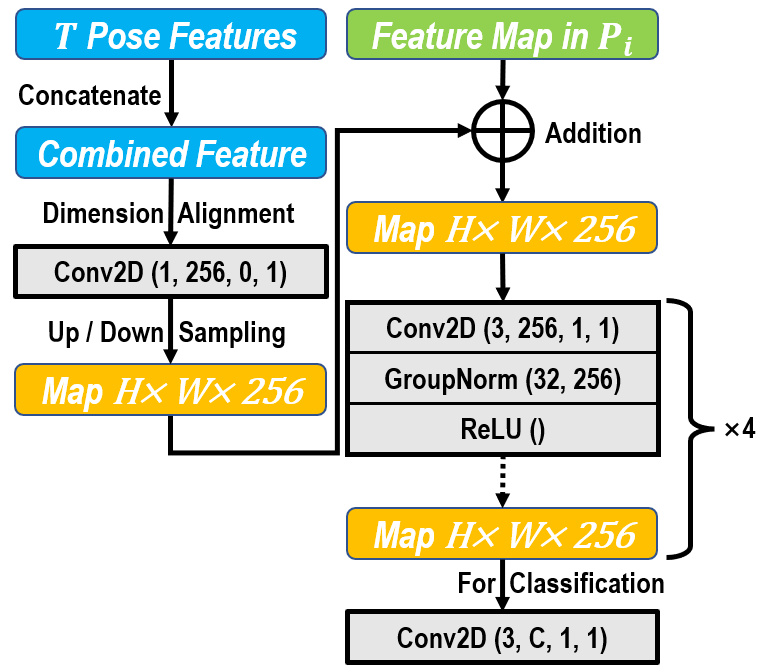}}
	\centerline{(a)}
	\end{minipage}
	\begin{minipage}{0.4\linewidth}
	\rightline{\includegraphics[width=1\textwidth]{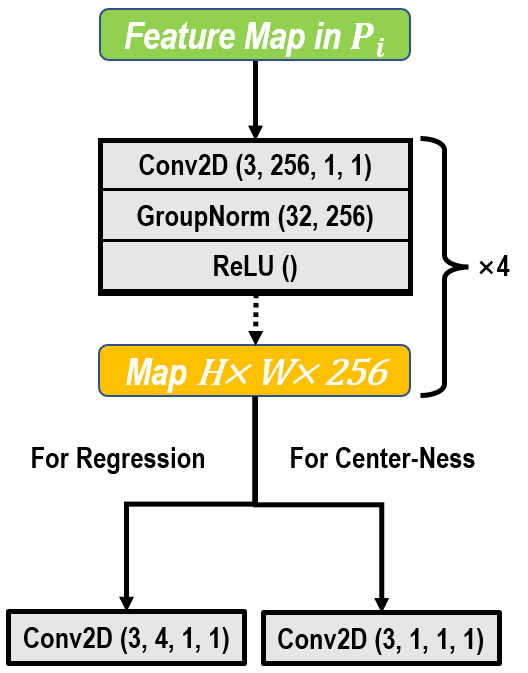}}
	\centerline{(b)}
	\end{minipage}
	\caption{Illustrations of feature extraction and aggregation in DangerDet. a) {\it Classification}. b) {\it Regression} and {\it Center-Ness} \cite{tian2019fcos}. Please note that the format of Conv2D is $(kernel\_size, out\_channels, padding, stride)$.}
	\label{fig3}
\end{figure}

In this part, we focus on the aggregation of multi-level features extracted from FPN and pose features predicted by stacked hourglasses. In the {\it Classification} prediction branch, as shown in Fig. \ref{fig3}(a), the input map is obtained by element-wise addition of two kinds of features. One is feature map in $\mathcal{P}_i$. The another one is transformed from multiple pose features by a series of sequential operations including concatenation, dimension alignment and up(down)-sampling. Then, we extract the high-level semantics through four $3\times3$ Conv2D layers with 256 filters. Layers GroupNorm and ReLU follow closely each convolution layer. Finally, through a $3\times3$ Conv2D layer with $C$ ($C=3$) filters, this branch outputs a result map with behavior category in pixel level. In Fig. \ref{fig3}(b), we maintain the same feature extraction structure as the classification branch. However, pose features are removed. The number of output channels of \emph{Regression} and \emph{Center-Ness} branched are changed into 4 and 1 accordingly. The outputs of these two branches are also heatmaps in pixel level. In the post-processing, these result heatmaps will be decoded to obtain behavior types and bounding-boxes.

\subsection{Loss Function}

Our DangerDet has two major losses: (i) For object detection part, the losses are $L_{cls}$, $L_{reg}$, and $L_{cen}$ for \emph{Classification}, \emph{Regression} and \emph{Center-Ness} respectively. (ii) For pose estimation part, the loss is the MSE for keypoints heatmap named $L_{kpt}$. The  final training loss function is defined as Eqn. \ref{eqn2}:

\begin{equation}\small
    \begin{aligned}
    L(\{p_{x,y}\}, \{t_{x,y}\}) =  \frac 1{N_{pos}}\sum_{x,y}{L_{cls}(p_{x,y}, c^*_{x,y})} +  \\  
    \frac \lambda{N_{pos}} \sum_{x,y}{1_{c^*_{x,y}>0}L_{reg}(t_{x,y}, t^*_{x,y})} + \sum^T_{t=1}L_{kpt}(C^r_t, C^r)
    \label{eqn2}
    \end{aligned}
\end{equation}
where $L_{cls}$ is focal loss \cite{lin2017focal} and $L_{reg}$ is the IOU loss as in UnitBox \cite{yu2016unitbox}. $N_{pos}$ denotes the number of positive samples. $\lambda$ is set to 1. $C^r_t$ in $L_{kpt}$ is the predicted joints confidence map. $C^r$ is the ground-truth keypoint heatmap. $T$ is the number of hourglass modules which is set as 4.

\section{Experiments}

In this section, we will describe the construction process of our danger behavior dataset, and then report the results of ablation experiments on this dataset.

\subsection{Our Danger Behavior Dataset}

Our raw data is collected from a primary school in Shanghai, China. It consists of about 100 surveillance videos captured by six 4K $(3840 \times 2160)$ cameras installed in a corridor without overlapping. These cameras start recording student activities during a pre-set period of time, about ten minutes after class. Fig. \ref{fig1}(a) gives an example frame. Then we sampled these videos every 0.5 seconds (skipping about 15 frames with FPS=30) and produced latent keyframes of our danger behavior dataset. After removing irrelevant monitoring areas, we cropped all the frames to $2400 \times 1600 (3:2)$. We used LabelImg \cite{tzutalin6labelimg}, a popular open-source annotation tool in object detection, to finish the annotation work.

Specifically, we selected three representative dangerous behavior categories to label. They are {\bf fight}, {\bf tumble}, and {\bf squat}. These behaviors are the precursor of potential danger, or themselves are dangerous actions, which need early warning in time once happening. Some examples of annotations are shown in Fig. \ref{fig1}(b)-(d). In addition, we have directly applied PifPaf \cite{kreiss2019pifpaf} to estimate pose of all annotated frames and generate rough keypoints coordinates of all students in our dataset. Finally, the dataset includes {\bf 7161} static images, and the number of three behaviors (fight, tumble, squat) is {\bf 665}, {\bf 6962}, and {\bf 1788}, respectively.

\subsection{Ablation Experiments}

\begin{table}[]\small
    \caption{Comparison between DangerDet and other methods.}
    \centering
    \begin{tabular}{c|ccc|ccc}
        \Xhline{1.5pt}
        {Method} & {resnet} & {pose} & {T} & $AP_{50}$  & $AP_{75}$  & {mAP}  \\
        \Xhline{1.5pt}
        FCOS & 50 & \XSolidBrush & 0 & 83.7 & 69.5 & 59.5  \\
        DangerDet & 50 & \Checkmark & 4 & 89.6 & 81.3 & 69.2  \\
        \hline
        FCOS* & 101 & \XSolidBrush & 0 & 89.4 & 78.4 & 67.6  \\
        DangerDet* & 101 & \Checkmark & 4 & 85.3 & 82.6 & {\bf 71.0}  \\
        \hline
        DangerDet1 & 50 & \Checkmark & 1 & 86.1 & 69.2 & 60.5  \\
        DangerDet2 & 50 & \Checkmark & 2 & 87.8 & 72.6 & 62.9  \\
        \Xhline{1.5pt}
    \end{tabular}
    \label{tab1}
\end{table}

We divided our dataset into training and validation set in the ratio 8:2. On the validation set, the DangerDet was evaluated by measuring precision, recall, and mean average precision (mAP) at an intersection-over-union (IOU) threshold from 0.5 to 0.95. For comparison, we trained the original FCOS model as the baseline without the pose module or $L_{kpt}$ part in loss function. The selected backbones include ResNet-50 and ResNet-101. Besides, we investigated the influence of different number $T$ of stacked hourglass modules in DangerDet.

\begin{table}[]\small
    \caption{Three behaviors' detection results of DangerDet*.}
    \centering
    \begin{tabular}{c|ccc}
        \Xhline{1.5pt}
        Behavior & fight & tumble & squat \\
        \Xhline{1.5pt}
        Labels & 132 & 1385 & 362 \\
        \hline
        mAP & {\bf 68.0} & {\bf 71.9} & {\bf 73.1}  \\
        \Xhline{1.5pt}
    \end{tabular}
    \label{tab2}
\end{table}

\begin{figure}[]
	\centering
	\includegraphics[width=0.48\textwidth]{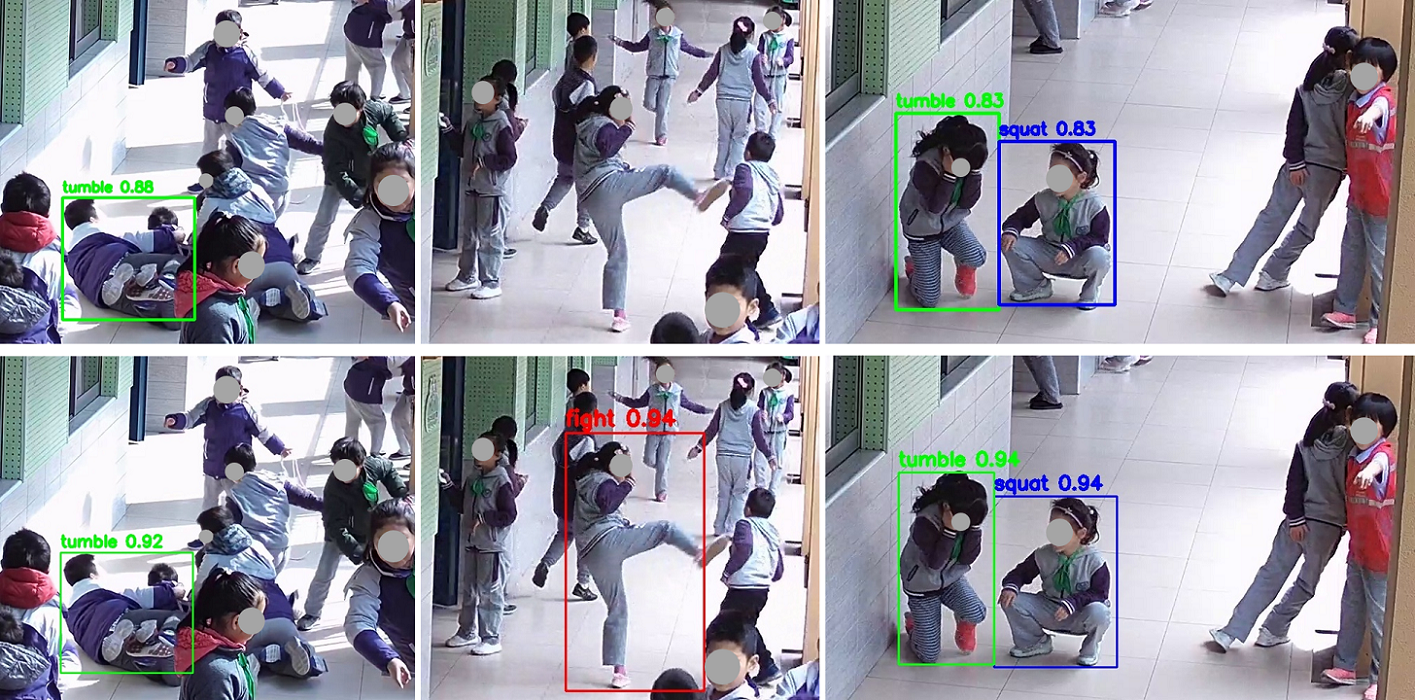}
	\caption{Detection results of baseline FCOS (top column) and our DangerDet (bottom column) with IoU threshold = 0.8.}
	\label{fig4}
\end{figure}

The validation results of DangerDet and other methods for comparison are shown in the table \ref{tab1}. With ResNet-50 and ResNet-101 as the backbone, our methods DangerDet(*) are 9.7\% and 3.4\% points higher than the baselines FCOS(*) in mAP, respectively. With ResNet-50 as the backbone, our methods DangerDet/1/2 with $T=1$, $T=2$ and $T=4$ always perform better than the baseline FCOS. With lighter backbone, DangerDet1 obtains larger mAP than FCOS*. All these proved the superiority of our method. With $T=4$ and the ResNet-101 as backbone, DangerDet* achieved the best result. Some qualitative results are shown in Fig. ~\ref{fig4}.

Table \ref{tab2} shows the detection results of three behaviors using DangerDet*. Unsurprisingly, the action fight gets worse accuracy than the other two behaviors for its higher dependency on continuous spatiotemporal features. We expect that this situation might be improved by collecting more data or associating continuous frames.

\section{Conclusion}

We presented a novel method DangerDet which integrates both object detection and pose estimation for detecting three dangerous behaviors of students on campus. The proposed end-to-end network structure allows us to easily combine deep features and two-dimensional keypoints. By designing efficient feature aggregation methods and adjusting the loss function, the training robustness and final accuracy of DangerDet are improved. The integration of these strategies achieves an impressive result in dangerous behaviors detection of the real school scenario. Our method has potential value to be applied to campus security alarm system. We have released our codes for academic use in \url{https://github.com/hnuzhy/fcos_pose}.

\bibliographystyle{IEEEbib}
\bibliography{refs}

\end{document}